\documentclass{article}
\usepackage{spconf,amsmath,graphicx}

\usepackage{booktabs}
\usepackage{graphicx}
\usepackage{tabularx}
\usepackage{multirow}
\usepackage{url}
\usepackage{array}
\usepackage{mdframed}
\usepackage{tcolorbox}
\usepackage{bm}
\usepackage{amsmath}
\usepackage{amssymb}
\usepackage{textcomp}
\usepackage{hyperref}
\usepackage{cleveref}

\Crefname{section}{Sec.}{Secs.}
\Crefname{table}{Tab.}{Tabs.}
\Crefname{figure}{Fig.}{Figs.}
\Crefname{equation}{Eq.}{Eqs.}
\Crefname{appendix}{Apx.}{Apx.}

\newcommand{\bz}{\bm{z}}

\newcommand{\bs}{\bm{s}}

\newcommand{\be}{\bm{e}}

\newcommand{\bvartheta}{\bm{\vartheta}}

\title{Explore the Vulnerability of Black Box Models via Diffusion Models}
\name{\textbf{Jiacheng Shi, Yanfu Zhang, Huajie Shao, Ashley Gao}}
\address{\small
  \begin{tabular}{c}
  Department of Computer Science, William \& Mary, USA\\
  \texttt{\{jshi12, yzhang105, hshao, ygao18\}@wm.edu}
  \end{tabular}
}
\begin{document}
%\ninept
%
\maketitle
\begin{abstract}
Recent advancements in diffusion models have enabled high-fidelity and photorealistic image generation across diverse applications. However, these models also present security and privacy risks, including copyright violations, sensitive information leakage, and the creation of harmful or offensive content that could be exploited maliciously. In this study, we uncover a novel security threat where an attacker leverages diffusion model APIs to generate synthetic images, which are then used to train a high-performing substitute model. This enables the attacker to execute \textbf{model extraction} and \textbf{transfer-based adversarial attacks} on black-box classification models with minimal queries, without needing access to the original training data. The generated images are sufficiently high-resolution and diverse to train a substitute model whose outputs closely match those of the target model. Across the seven benchmarks, including CIFAR and ImageNet subsets, our method shows an average improvement of 27.37\% over state-of-the-art methods while using just $0.01\times$ of the query budget, achieving a 98.68\% success rate in adversarial attacks on the target model. The code and supplementary materials are available on \href{https://github.com/jiachengQAQ/Attacking-Black-Box-Models}{this link}.

\end{abstract}

\section{Introduction}
\label{sec:intro}

Black box machine learning models are susceptible to various attacks, including model extraction~\cite{truong2021data,rosenthal2023disguide} and adversarial transfer attacks~\cite{zhou2020dast,zhang2022towards}. In a model extraction attack, an attacker attempts to replicate a target model's functionality by querying it and using the responses to reconstruct a substitute model. On the other hand, an adversarial transfer attack involves generating adversarial examples to attack one model (substitute model) and, because the substitute model is well-trained to approximate the target model, an attacker can successfully use those adversarial examples to attack the target model as well ~\cite{demontis2019adversarial}. Both attacks ultimately aim to create a substitute model to facilitate these exploits.

Previously, there were two methods (GANs and VAEs) to obtain a substitute model based on the fact that an attacker can use generated photorealistic images to query the target model and gather information about its decision-making process. State-of-the-art image generation includes Generative Adversarial Networks~\cite{thanh2020catastrophic} (GANs) and Variational Autoencoders~\cite{sun2021adversarial}(VAEs). However, previous research has shown that training GANs presents significant challenges. For instance, studies~\cite{ thanh2020catastrophic} have highlighted issues such as model collapse and convergence difficulties, which must be overcome to achieve a robust image generator, in addition to requiring substantial machine learning engineering efforts for parameter optimization. Similarly, state-of-the-art Variational Autoencoder (VAE)-based approaches~\cite{sun2021adversarial} can also generate images. However, the assumption of a simple (often Gaussian) prior distribution in the latent space may limit the expressiveness of VAEs. This assumption might not adequately capture the true complexity of the data distribution, resulting in suboptimal generative performance.

Rapid advancements in diffusion models enable high-quality, photo-realistic image generation by leveraging large-scale, diverse datasets.
%~\cite{song2019generative}. 
These images can train substitute models to attack black-box classification models through model extraction and adversarial transfer attacks. Unlike GAN- and VAE-based methods, diffusion models offer superior generation capabilities, making them well-suited for such attacks. Their integration with APIs enhances accessibility, scalability, and usability, solidifying their appeal for exploiting black-box classification models. The recent study~\cite{shao2023data} focuses exclusively on adversarial transfer attacks, leveraging synthetic images from diffusion models to train substitutes. However, it heavily relies on extensive target model queries for Membership Inference, resulting in high query budgets even under relaxed $L_2$ norms, making it impractical.
%~\shao{discuss related work on diffusion based attacks, the reviewers mentioned one paper in this topic}

Our study explores how diffusion model APIs can be misused to attack black-box classification models by obtaining substitute models. Model extraction and adversarial transfer attacks typically require difficult-to-access information like gradients and logits. However, our approach eliminates this need, as attackers can exploit generative model APIs (e.g., Stable Diffusion, DALLE-3, Imagen) by inputting benign prompts to generate high-quality synthetic data. This data trains substitute models, enabling efficient model extraction and adversarial attacks on black-box classification models with minimal queries, no real or private data, and lower attack deployment costs.

%Motivated by the advantages offered by diffusion APIs, our study proposes that diffusion models' APIs can be (mis)used to attack black box classification models by obtaining a substitute model. Achieving successful model extraction and adversarial transfer attacks on black-box classification models remains challenging, as it requires access to difficult-to-obtain information, such as gradient data and logits. In our study, the attackers do not need gradient information or jailbreak the black-box classification model, thus effectively solving the challenges (that such attackers do not require any gradient information or logits output from the target model). Instead, we can exploit this vulnerability by simply misusing the API of generative models (e.g., Stable Diffusion~\cite{rombach2022high}, DALLE-3~\cite{betker2023improving} Imagen~\cite{saharia2022photorealistic}), inputting benign prompts to obtain high-quality synthetic data. This synthetic data can then be used to train substitute models, enabling attackers to perform model extraction and adversarial attacks on online black-box classification models with minimal query iterations, without the need for any real or private data. This method not only streamlines the attack deployment process but also significantly enhances attack efficiency and lower the query budget.

%

% no finding. Previously laid down on things work. API is convenient. Motivated by this, we propose our API-based model

\begin{figure}[!t]
  \centering
  \vspace{-1mm}
  \includegraphics[scale=0.35]{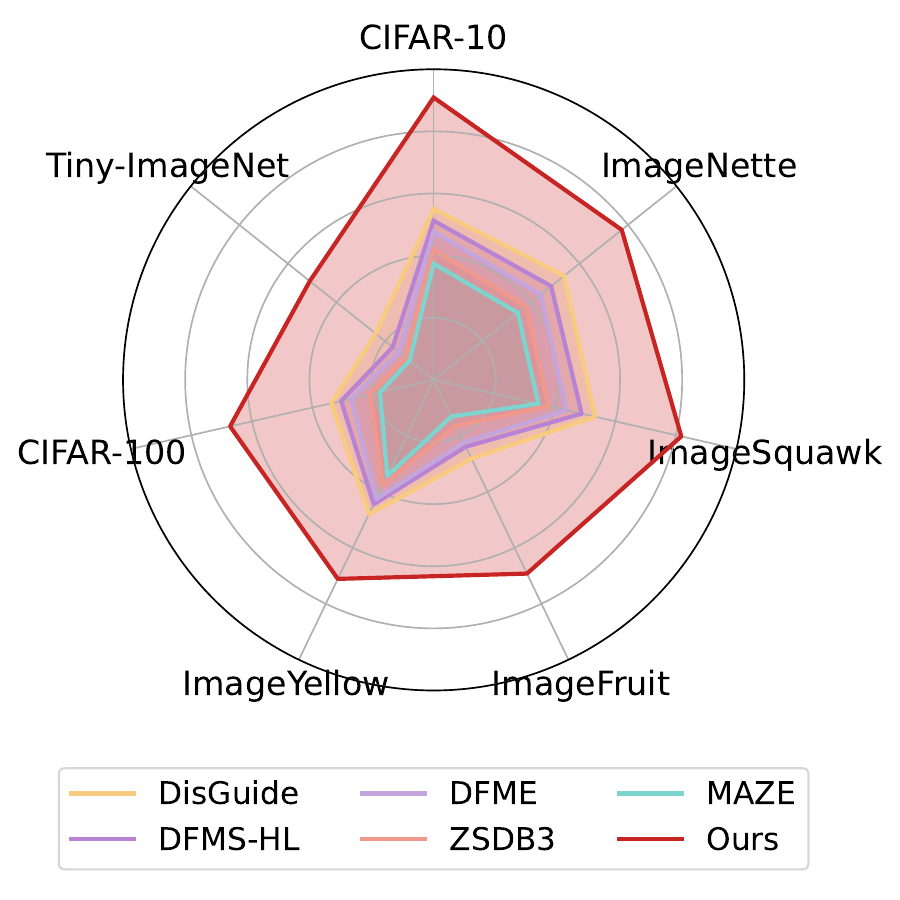}
  \vspace{-2mm}
  \caption{Previous methods struggle to handle limited query budgets. To ensure a fair comparison, we scale up the baseline query count by 100 times across seven datasets, and our method still improved the test accuracy by 21.4\% over the best prior baseline on the ImageNette dataset.}
  \label{fig:rada}
  \vspace{-6mm}
\end{figure}

We demonstrate the efficiency and generalizability (i.e. our method works on different black box classification model architectures) of our method across seven benchmarks, achieving an average improvement of 27.37\% over state-of-the-art methods while using just 0.01x of the query budget. Specifically, on the Tiny-ImageNet and CIFAR-100 datasets, our approach outperforms previous baselines by 24.4\% and 30.1\%, respectively, while requiring only 0.01x of the query budget, as shown in~\Cref{fig:rada}. Moreover, our method achieves an impressive 81.8\% and 79.19\% top-one accuracy on the CIFAR-10 and ImageNette datasets with just $5k$ and $1k$ queries, respectively, as shown in~\Cref{tab:t1_hard}, establishing a new state-of-the-art. %These results underscore the practicality of our approach in real-world scenarios.

In summary, our main contributions include:
\begin{itemize}
%\item We identify a \textbf{new security vulnerability by exploiting the image generation capabilities of generative model APIs}, implicitly threatening the security of sophisticated black-box models. Our research shows that attackers can readily leverage online diffusion models APIs,utilizing a benign prompt strategy to generate high-resolution and diverse synthetic images. These images are subsequently used to train substitute models that are equipped to execute both model extraction and adversarial transfer attacks. Such attacks, which are straightforward to deploy yet difficult to defend against, pose a significant threat to the security of black-box systems in practical scenarios.
\item We identify \textbf{a new security vulnerability in sophisticated black-box models.} Our research shows that attackers can readily leverage online diffusion models APIs,
utilizing a benign prompt strategy to generate high-resolution and diverse synthetic images. These images are subsequently used to train substitute models that are equipped to execute both model extraction and adversarial transfer attacks. Such attacks, which are straightforward to deploy yet difficult to defend against, pose a significant threat to the security of black-box systems in practical scenarios.
\item We propose an \textbf{API-based solution that significantly reduces the query budget} required for deploying model extraction and adversarial transfer attacks against black-box systems, and through experiments, we demonstrate its efficiency and economic viability with minimal query requirements.
\item We conduct \textbf{extensive experiments across various real-world datasets} to validate the effectiveness and feasibility of our method, which does not utilize any real data. Specifically, on the CIFAR-10 and ImageNet subsets, our approach increases the top-one accuracy of the best baseline model extraction method by 32.3\% and 21.4\%, respectively, while utilizing only 1\% of the baseline query volume, as shown in~\Cref{fig:rada}. Furthermore, our method significantly enhances the success rates of adversarial transfer attacks, achieving increases of 41.8\% on CIFAR-10 and 58.2\% on CIFAR-100 compared to the best baseline.
\end{itemize}

\section{Related Work}
%Our work builds on advancements in diffusion models~\cite{schramowski2023safe}, data-free model extraction~\cite{truong2021data}, and adversarial attacks using substitute models~\cite{zhou2020dast}. \textbf{A comprehensive discussion of these works is provided in Appendix~\ref{sec:related_work}}.

\textbf{Diffusion Model}. Diffusion-based generative models, known for generating high-fidelity and diverse synthetic images, are central to our research, particularly stable and latent diffusion models. While these models are powerful, their security and privacy implications warrant critical examination. Previous studies~\cite{schramowski2023safe} address security concerns, proposing methods to mitigate risks from inappropriate inputs. However, the rapid evolution of diffusion technologies introduces challenges, such as generating Not-Safe-for-Work (NSFW) content, leading to biases, copyright issues, and harmful content~\cite{schramowski2023safe}. Moving beyond these existing issues, our work identifies the potential misuse of generative models for training substitute models, which can then be used for model extraction and adversarial transfer attacks, posing significant risks to black-box systems.

% \textbf{Data Free Model Extraction}. The rise of Machine Learning Models as a Service (MLaaS) leads to proprietary models offering inference through cloud-based APIs like Google Vision AI\footnote{\url{https://cloud.google.com/vision}}, making them vulnerable to adversarial model extraction attacks. Traditional knowledge distillation methods~\cite{romero2014fitnets}, requiring white-box access and genuine training data, are unsuitable for black-box MLaaS models. Black-box approaches like MAZE~\cite{kariyappa2021maze}, DFME~\cite{truong2021data}, and ZSDB3~\cite{wang2021zero} use GANs to generate training data but face challenges such as model collapse and high query costs, particularly in soft-label settings (i.e., probabilities or logits). More recent methods like DFMS~\cite{sanyal2022towards} and DisGuide~\cite{rosenthal2023disguide}, designed for hard-label scenarios (i.e., return top-1 prediction), better align with real-world applications but still require substantial query budgets. SHAP employs energy-based loss to enhance query efficiency, while DualCOS leverages a dual clone architecture to improve model extraction performance; however, both methods share the limitation of requiring a query budget in the millions. Our work addresses these challenges by training a robust substitute model with significantly fewer queries, thereby exposing security vulnerabilities in MLaaS.
\noindent\textbf{Data-Free Model Extraction}. The proliferation of Machine Learning Models as a Service (MLaaS) has made proprietary models accessible through cloud-based APIs like Google Vision AI\footnote{\url{https://cloud.google.com/vision}}, rendering them vulnerable to model extraction attacks. Traditional knowledge distillation methods~\cite{romero2014fitnets}, requiring white-box access and genuine training data, are ineffective for black-box MLaaS models. Black-box approaches such as MAZE~\cite{kariyappa2021maze}, DFME~\cite{truong2021data}, and ZSDB3~\cite{wang2021zero} synthesize training data using GANs but face challenges like mode collapse and high query costs, particularly in soft-label settings. Recent methods, including DFMS~\cite{sanyal2022towards} and DisGuide~\cite{rosenthal2023disguide}, focus on hard-label scenarios, aligning better with real-world applications but still requiring significant query budgets. SHAP~\cite{kumar2024vidmodex} improves query efficiency through energy-based loss, and DualCOS~\cite{yang2024dualcos} enhances model extraction using a dual clone architecture, yet both demand millions of queries. In contrast, our work trains a robust substitute model with far fewer queries, exposing critical vulnerabilities in MLaaS systems.

\noindent\textbf{Data-Free Adversarial Attack with Substitute Models}.
\label{sec:related_dfta}The rise of deep learning has made commercial models vulnerable to parameter exfiltration through `pay-per-query' systems. Early methods like JPBA~\cite{papernot2017practical} and Knockoff~\cite{orekondy2019knockoff} train substitute models using synthetic data labeled by target models, limiting their applicability to truly data-free scenarios. Recent studies, such as DaST~\cite{zhou2020dast} and TEBA~\cite{zhang2022towards}, optimize label control loss or query volume but still rely on significant interaction with the target model. DifAttack~\cite{liu2024difattack} enhances performance with disentangled features but depends on real data during training, compromising user privacy. while, DERD~\cite{zhou2024derd} leverages dual-level adversarial learning for data-free robustness. However, these approaches remain economically infeasible due to high query demands. To address these challenges, our method employs online diffusion model APIs to generate high-quality synthetic images, significantly reducing query budgets while improving surrogate model effectiveness.

\section{Proposed Method}
\subsection{Overview}
\begin{figure*}[!t]
  \centering
  \vspace{-5mm}
  \includegraphics[scale=0.41]{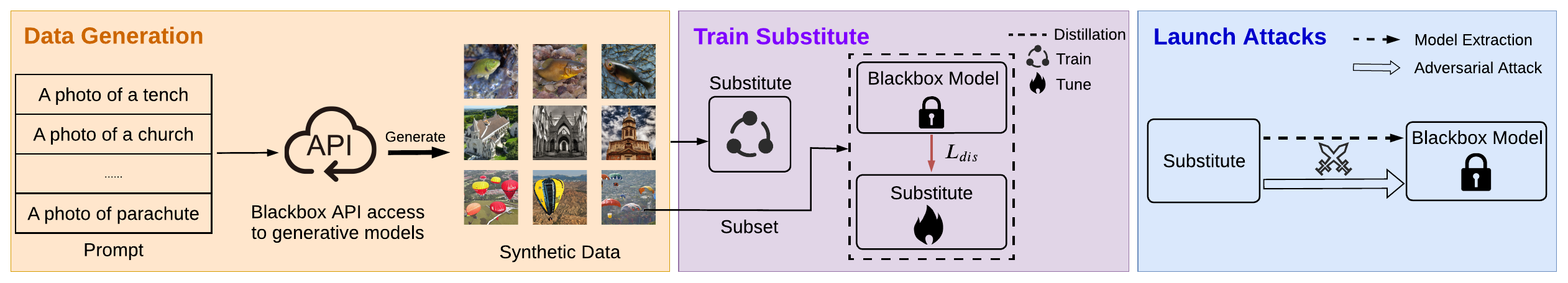}
  \caption{ Illustration of the API-based attack solution, consisting of three stages: Data Generation, Train Substitute Model, and Launch Attacks. (a) Data generation aims to produce high-quality and diverse synthetic images. (b) Train Substitute Model stage pre-trains a substitute model with synthetic data and fine-tunes it through knowledge distillation from the black-box model. (c) Launch Attacks stage deploys the trained substitute model to execute model extraction and adversarial attacks on the black-box model.
  }
  %\vspace{-1mm}
  \label{fig:pipline}
  \vspace{-4mm}
\end{figure*}

The objective of our approach is to utilize off-the-shelf generative models to synthesize high-quality images and efficiently train robust substitute models for deploying two types of attacks. The entire framework is divided into three parts: (i) In the data generation phase as shown in~\Cref{fig:pipline}, we transition from traditional synthesizers, which require extensive queries to the target model, to employing online stable diffusion APIs for generating high-quality images. This approach achieves zero-query generation, substantially reducing the query budget typically associated with the production of synthetic data. (ii) In the substitute model training phase, we deviate from conventional methods that often face convergence issues when starting from scratch. By leveraging synthetic images produced in the initial stage, we pre-train the substitute model, thus establishing a robust knowledge base. The model distillation phase then focuses on fine-tuning the substitute model with limited queries to ensure that its outputs are closely aligned with those of the target model. (iii) Finally, we launch model extraction and adversarial attacks to challenge the security of black-box models. we will discuss them in the following sections.
%This streamlined methodology not only enhances practical feasibility and economic efficiency but also poses a significant challenge to the security of black-box systems in operational environments. 

\subsection{Data Generation}
The online Stable Diffusion APIs are at the forefront of generating high-quality images. We utilize these APIs to generate synthetic images for pre-training the substitute model. By default, we employ class-level prompts, introducing class names $C = {c_1, \dots, c_k} = \{"tench", "church", \dots, "parachute"\}$ with prompts $p_i$ = "a photo of a {$c_i$}", which are input into the APIs. For the Stable Diffusion models deployed within APIs, the mechanism to infer images based on prompts operates as follows: The prompt $p_i$ is passed through an encoder (e.g., from CLIP) to generate an embedding $\be_i$, which conditions the noise vector $\bz_i$. The synthetic image $\bs_i$ is produced by applying the Stable Diffusion model $G$ to the noise vector $\bz_i$ guided by the embedding $\be_i$ as follows:
\begin{equation}
\bs_i = G(\bz_i, \be_i) = \sqrt{\beta} \sum_{t=1}^{T} \sqrt{1-\beta^t} \cdot \frac{1}{\sqrt{T}} \cdot G_{\bvartheta_t}(\bz_i, \be_i),
\end{equation}
where $\bz_i$ represents the random noise, while $G_{\bvartheta_t}$ is the denoising network at step $t$, parameterized by $\bvartheta_t$. The factor $\beta$ regulates the balance between image fidelity and variation, with $T$ indicating the total number of steps in the diffusion process. The generated image is iteratively refined by denoising at each step until the final output $\bs_i$ is obtained. All training parameters in SD are set as default, following the configurations in~\cite{rombach2022high}.
\subsection{Acquisition of Substitute Models}
\label{sec:acq_sub}
To address the impracticality of high query demands in model extraction, we depart from traditional min-max approaches~\cite{ truong2021data} and focus on efficiently training a robust substitute model. By leveraging pre-training on a large-scale, high-quality synthetic dataset, our method significantly reduces the query budget for model extraction and adversarial transfer attacks. In the Train Substitute stage (\Cref{fig:pipline}), the substitute model is pre-trained on diverse synthetic data generated in the initial phase, providing a robust foundation for subsequent knowledge distillation. A subset of these images is selected with equal probability across categories and queried once to obtain hard labels from the target model $T$, denoted as $y_T = T(\hat{x})$. These labels are treated as ground truth for training the substitute model $S$, parameterized by $\theta_S$, to minimize the cross-entropy loss $L_{dis} = CE(S(\hat{x}, \theta_S), y_T)$. This single-query strategy ensures economic feasibility by significantly reducing interactions with the target model compared to prior methods. Upon training, the substitute model achieves predictions closely aligned with the target model, enabling effective model extraction and adversarial attacks with minimal queries.

\subsection{Substitute Model for Adversarial Transfer Attack}
In the process of attacking the substitute model, we employ white-box methods such as BIM, FGSM, and PGD. These strategies exploit the model's gradients to iteratively construct perturbations, $\delta$, bound by a specified limit, $\epsilon$, ensuring that the adversarial samples, $x_{adv}$, remain imperceptible to human observers. Our training parameters follow the settings in~\cite{zhang2022towards}. For each attack, the perturbation $\delta$ is chosen to maximize the increase in the loss function, $L$, given the input $x$ and label $y$, as articulated by the equation $x_{adv} = x + \delta$, where $\delta = \arg\max_{|\delta| \leq \epsilon} L(f(x + \delta), y)$. By introducing these calculated disturbances into the substitute model, we observe and quantify the effect of these incremental adversarial changes on the model's gradients and loss, thereby identifying the optimal adversarial perturbation to challenge the model's robustness. Our primary objective is to develop a method for conducting transfer attacks effectively, thereby highlighting the practicality of such security concerns in real-world contexts.

\section{Experiments}

\subsection{Experiments on Model Extraction} 
\label{sec:dfme}

\begin{table*}[!t]
\centering
\caption{ Accuracy (\%) of substitute models on datasets with CIFAR-10, and ImageNet subsets in the hard-label setting with various target models. We use ResNet-18
as the default substitute network. All results are averaged over three random seeds.}
\label{tab:t1_hard}
%\vspace{-1mm}
\scalebox{0.64}{
%\resizebox{\textwidth}{!}{
\begin{tabular}{c c c c c c c c c c}
\toprule
\small Dataset  & \small Model & \small Target & \small MAZE~\cite{kariyappa2021maze} & \small DFME~\cite{truong2021data} & \small ZSDB3~\cite{wang2021zero}  & \small DFMS-H~\cite{sanyal2022towards}L & \small DisGuide~\cite{rosenthal2023disguide}  & \small Ours & \small Query Budget\\
\midrule
\small \multirow{3}{*}{ {\small{}CIFAR-10}} & \small AlexNet & \small 84.93 & \small 10.97 & \small 10.84 & \small 10.38 & \small 11.89 & \small 11.62  & \small \textbf{73.88} & \\
 & \small VGG-16 & \small 91.41 & \small 10.52 & \small 10.96 & \small 10.43 & \small 10.71 & \small 12.49  & \small \textbf{79.19} & \\
 & \small ResNet-34 & \small 93.92 & \small 10.55 & \small 10.51  & \small 10.83 & \small 11.17  & \small 12.57  & \small \textbf{81.51} & \small \multirow{-3}{*}{$5k$}\\
 
 \midrule
\small \multirow{3}{*}{ImageNette} & \small VGG-16 & \small 91.69 & \small 10.38 & \small 10.61  & \small 10.54 & \small 10.83  & \small 11.31 & \small \textbf{67.95} &  \\
 & \small WRN16 & \small 91.55 & \small 10.89 & \small 10.52 & \small 10.98 & \small 11.44 & \small 12.72  & \small \textbf{63.52} & \\
 & \small ResNet-34 & \small 92.26 & \small 10.88 & \small 10.31  & \small 10.56 & \small 10.82  & \small 11.31   & \small \textbf{69.97} & \small \multirow{-3}{*}{$1k$} \\
 \midrule
\small \multirow{3}{*}{ImageSquawk} & \small VGG-16 & \small 91.82 & \small 10.23 & \small 10.86 & \small 10.24 & \small 10.88 & \small 12.93  & \small \textbf{70.33} &  \\
 & \small WRN16 & \small 92.25 & \small 10.19 & \small 10.53 & \small 10.79 & \small 10.74 & \small 11.85  & \small \textbf{71.52} & \\
 & \small ResNet-34 & \small 92.47 & \small 10.44 & \small 10.39  & \small 10.21 & \small 10.71  & \small 11.53   & \small \textbf{73.64} & \small \multirow{-3}{*}{$30$} \\
 \midrule
 \small \multirow{3}{*}{ImageFruit}     & \small VGG-16 & \small 80.25 & \small 10.35 & \small 10.16 & \small 10.83 & \small 10.55 & \small 11.03 & \small \textbf{65.81} &  \\
 & \small WRN16 & \small 79.04 & \small 10.49 & \small 10.39 & \small 10.35 & \small 10.92 & \small 11.43  & \small \textbf{63.22} &\\
 & \small ResNet-34 & \small 78.52 & \small 10.11 & \small 10.49   & \small 10.16 & \small 10.28  & \small 10.98   & \small \textbf{62.36} & \small \multirow{-3}{*}{$130$} \\
 \midrule
\small \multirow{3}{*}{ImageYellow} &  \small VGG-16 & \small 91.35 & \small 10.34 & \small 10.53 & \small 10.29 & \small 10.63 & \small 12.55  & \small \textbf{66.73} &  \\
 & \small WRN16 & \small 90.16 & \small 10.73 & \small 10.95 & \small 11.25 & \small 11.07 & \small 12.18  & \small \textbf{63.05} & \\
 & \small ResNet-34 & \small 90.82 & \small 10.26 & \small 10.45   & \small 10.66 & \small 11.24   & \small 11.59   & \small \textbf{64.14} & \small \multirow{-3}{*}{$50$} \\

 \bottomrule
\end{tabular}}
\vspace{-0.2cm}
\end{table*}

\begin{table*}[t]
\centering
\caption{Accuracy (\%) of student models on datasets of hundreds of classes in the hard-label setting with various target models. All results are averaged over three random seeds.}
\label{tab:t2_hard}
%\vspace{-2mm}
%\resizebox{\columnwidth}{!}{
\scalebox{0.61}{
\begin{tabular}{c c c c c c c c c c}
\toprule
\small Dataset  & \small Model & \small Target & \small MAZE~\cite{kariyappa2021maze} & \small DFME~\cite{truong2021data} & \small ZSDB3~\cite{wang2021zero}  & \small DFMS-H~\cite{sanyal2022towards}L & \small DisGuide~\cite{rosenthal2023disguide}  & \small Ours & \small Query Budget\\
\midrule
\small \multirow{2}{*}{CIFAR-100} 
 & \small VGG-16 & \small 69.66 & \small 1.07 & \small 1.06 & \small 1.03 & \small 1.05 & \small 1.13  & \small \textbf{57.35} & \\
 & \small ResNet-34 & \small 79.89 & \small 1.03 & \small 1.05  & \small 1.05  & \small 1.12 & \small 1.27  & \small \textbf{60.51} & \small \multirow{-2}{*}{$150k$}\\
 
 \midrule
\small \multirow{2}{*}{Tiny-ImageNet} & \small VGG-16 & \small 53.14 & \small 0.53 & \small 0.57 & \small 0.54 & \small 0.56 & \small 0.63 & \small \textbf{37.18} &  \\
 & \small ResNet-34 & \small 64.55 & \small 0.51 & \small 0.50  & \small 0.53  & \small 0.55 & \small 0.65  & \small \textbf{45.97} & \small \multirow{-2}{*}{$200k$} \\
 \bottomrule
\end{tabular}}

\vspace{-0.2cm}
\end{table*}

\subsubsection{Performance Comparison on Small Datasets}
%\jz{say sth like query budget is important, we first fix the query budget to a small value for all methods.}
Query budget is a critical factor in evaluating the efficiency of adversarial methods. To ensure a fair comparison, we initially fixed the query budget to a small value across all methods. As shown in~\Cref{tab:t1_hard}, our method consistently outperforms baselines, achieving 81.58\% accuracy on CIFAR-10 under a $5k$ query limit, compared to DisGuide's 12.57\%. On average, our approach surpasses the best baseline by 51.4\% across all datasets in hard-label settings and more training detial as show in Appendix~\ref{sec:dfme_setting}. Further evaluation with different target and substitute model architectures (\Cref{tab:diff_sub}) shows that our method improves baseline performance by at least 68\% on CIFAR-10 and 55\% on CIFAR-100. For a more stringent test, we fixed our query budget while allowing baselines up to $500k$ queries on CIFAR-10 and ImageNet (\Cref{fig:rada}). Even under this condition, our method maintained at least a 20.8\% performance advantage. In soft-label settings as shown in Appendix~\ref{sec:soft_me} , our approach outperformed baselines by 59.5\% on CIFAR-10, demonstrating its robustness. We also explored the impact of pre-trained models in Appendix~\ref{sec:abl}. Visualization of baseline methods under limited query budgets (\Cref{fig:vis_baseline}) highlights their inefficiencies, with test accuracy around 10\%, comparable to random guessing. This reflects fundamental issues such as dependence on min-max games requiring extensive black-box interactions and challenges like GAN convergence failures and model collapses, which severely limit baseline effectiveness in query-limited environments. We also demonstrate the efficiency of our method in Appendix~\ref{sec:dfme_prove}. Since SHAP~\cite{kumar2024vidmodex} and DualCOS~\cite{yang2024dualcos} lack query efficiency and released code, a direct comparison is currently infeasible.

\subsubsection{Performance Comparison on Datasets with Hundreds of Classes}
Our method consistently outperforms prior baselines on datasets with hundreds of classes (\Cref{tab:t2_hard}). On CIFAR-100, it achieves 60.51\% accuracy under a $150k$ query budget, significantly surpassing DisGuide~\cite{rosenthal2023disguide}, which reaches only 1.27\%. This represents a performance improvement of at least 36.55\% in the hard-label setting. Moreover, across different substitute model architectures, our approach achieves an average improvement of 56.43\% on CIFAR-100 (\Cref{tab:diff_sub}).

Even under more relaxed conditions, where baselines are allowed a query budget of 2 million while ours remains fixed, our method maintains a performance advantage of at least 25\% (\Cref{fig:rada}). In the soft-label setting as shown in Appendix~\ref{sec:soft_me}, our approach also demonstrates superior performance, outperforming previous baselines by 57.6\% on Tiny-ImageNet. More experiments with substitute models based on different architectures are provided in Appendix~\ref{sec:different_sub}.

\begin{figure}[htbp]
%\vspace{-4mm}
  \centering
  \includegraphics[scale=0.26]{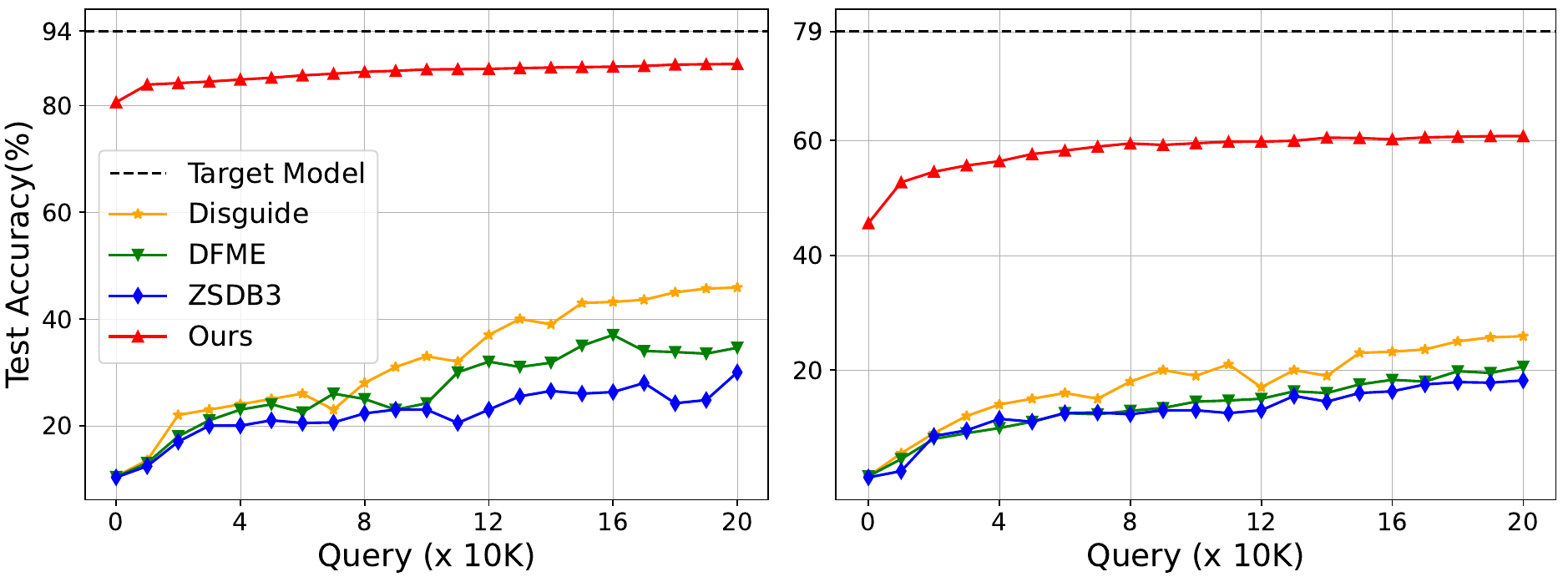}
  \vspace{-4mm}
  \caption{Evaluation of our method alongside other approaches (DisGuide, ZSDB, DFME) with query budget $200k$ on (\textbf{Left}) CIFAR-10 , and (\textbf{Right}) CIFAR-100. Across all query budget levels, our method improves baseline performance by an average of 51.6\% and 38.5\% on CIFAR-10 and CIFAR100.}
  \label{fig:query_baselines}
  \vspace{-4mm}
\end{figure}
\subsubsection{Performance Under Different Query Budget}

% In previous experiments, we demonstrate that our method could effectively extract knowledge from the target model on the CIFAR-10 dataset with a minimal query budget of $5k$ queries, achieving 87\% of the target model's test accuracy. We structure the experiments to gradually increase the query budget from 0 to $200k$ queries and compare our results with three previous baselines in data-free model extraction: DisGuide~\cite{rosenthal2023disguide}, DFME~\cite{truong2021data}, and ZSDB~\cite{wang2021zero}. 

% As shown on left of~\Cref{fig:query_baselines}, our model achieves a strong accuracy of 80.6\% on CIFAR-10 without any queries, thanks to extensive pre-training with synthesized data. As the query budget increases, the substitute model's test accuracy steadily approaches the target model's, reaching 87.86\% at $200k$ queries, close to the target model's 93\% accuracy.

% In further experiments depicted on the right side of~\Cref{fig:query_baselines} on the CIFAR-100 dataset, our model achieves a test accuracy of 45.5\% even with a zero query budget. Across all levels of query budgets, our method enhances the performance of the baselines by 38.5\%. These results underscore the efficiency and effectiveness of our approach in knowledge extraction and model performance, demonstrating significant improvements over existing baseline methods.
In previous experiments, our method effectively extracts knowledge from the target model on CIFAR-10, achieving 87\% accuracy with just $5k$ queries. We compared our results against three baselines: DisGuide, DFME, and ZSDB, across query budgets from 0 to $200k$. As shown on the left of~\Cref{fig:query_baselines}, our model reaches 80.6\% accuracy on CIFAR-10 without any queries, thanks to pre-training with synthesized data. With $200k$ queries, accuracy climbs to 87.86\%, nearing the target model's 93\%. On CIFAR-100 (right of~\Cref{fig:query_baselines}), our model achieves 45.5\% accuracy with zero queries, outperforming baselines by 38.5\% across all query budgets. %These results highlight the efficiency and superiority of our approach in model extraction.

\begin{table}[t]
\centering
\caption{ASR(\%) comparisons between our proposed method and baselines over CIFAR-10 and CIFAR-100 under a hard-label setting with a query budget Q = $150k$. All results are averaged over three random seeds.} 
\scalebox{0.6}{
\begin{tabular}{c c ccc ccc}
\toprule
\multirow{2}{*}{{Dataset}} & Type &
\multicolumn{3}{c}{ \small{}Targeted, hard-label}  & \multicolumn{3}{c}{ \small{}Untargeted, hard-label} \tabularnewline
\cmidrule(lr){2-8}
 & {\small{}Method} & {\small{}FGSM} & {\small{}BIM} & {\small{}PGD} & {\small{}FGSM} & {\small{}BIM} & {\small{}PGD} \tabularnewline

\midrule 

\multirow{6}{*}{{\small{}CIFAR-10}}
& {\small{}JPBA~\cite{papernot2017practical}} & {\small 2.13} & {\small 3.19} & {\small 3.37} & {\small 7.48} & {\small 10.76} & {\small 7.79} \tabularnewline

& {\small{}Knockoff~\cite{orekondy2019knockoff}} & {\small 1.98} & {\small 3.05} & {\small 3.09} & {\small 6.31} & {\small 9.47} & {\small 7.05} \tabularnewline

& {\small{}DaST~\cite{zhou2020dast}} & {\small 2.37} & {\small 3.71} & {\small 3.64} & {\small 8.62} & {\small 11.94} & {\small 8.05} \tabularnewline

& {\small{}DFME~\cite{truong2021data}} & {\small 3.71} & {\small 10.95} & {\small 8.06} & {\small 14.78} & {\small 19.54} & {\small 17.75} \tabularnewline

& {\small{}TEBA~\cite{zhang2022towards}} & {\small 7.05} & {\small 29.76} & {\small 22.5} & {\small 33.17} & {\small 55.43} & {\small 49.44} \tabularnewline

\cmidrule(lr){2-8}
& {\small{}\textbf{Ours}} & {\small \textbf{15.78}} & {\small \textbf{70.72}} & {\small \textbf{66.87}} & {\small \textbf{63.54}} & {\small \textbf{70.79}} & {\small \textbf{66.78}} \tabularnewline

\midrule 

\multirow{6}{*}{{\small{}CIFAR-100}}
& {\small{}JPBA~\cite{papernot2017practical} } & {\small 2.48} & {\small 3.29} & {\small 3.52} & {\small 8.92} & {\small 11.55} & {\small 8.85} \tabularnewline

& {\small{}Knockoff~\cite{orekondy2019knockoff} } & {\small 2.19} & {\small 3.12} & {\small 3.22} & {\small 7.33} & {\small 10.72} & {\small 8.12} \tabularnewline

& {\small{}DaST~\cite{zhou2020dast}} & {\small 3.25} & {\small 4.15} & {\small 4.28} & {\small 9.83} & {\small 13.22} & {\small 9.52} \tabularnewline

& {\small{}DFME~\cite{truong2021data}} & {\small 4.12} & {\small 11.25} & {\small 9.02} & {\small 15.78} & {\small 19.65} & {\small 17.43} \tabularnewline

& {\small{}TEBA~\cite{zhang2022towards}} & {\small 7.89} & {\small 28.56} & {\small 22.87} & {\small 32.45} & {\small 53.89} & {\small 47.95} \tabularnewline

\cmidrule(lr){2-8}
& {\small{}\textbf{Ours}} & {\small \textbf{16.24}} & {\small \textbf{68.97}} & {\small \textbf{64.85}} & {\small \textbf{60.45}} & {\small \textbf{68.34}} & {\small \textbf{65.76}} \tabularnewline

\bottomrule
                    \end{tabular}}
\label{tab:asr-hard-label}
\vspace{-8mm}
\end{table}
\subsection{Experiments on Adversarial Transfer Attack}
\label{sec:dfta}
We present the attack success rate (ASR) across various scenarios, differentiating between targeted and untargeted strategies, as well as hard-label and soft-label settings, using BIM, FGSM, and PGD attacks. For CIFAR-10 and CIFAR-100, we set perturbation limit $\epsilon$ = 8/255 and the step size at $\alpha$ = 2/255, with additional evaluation details provided in (Appendix~\ref{sec:dfta_setting}). According to the results in~\Cref{tab:asr-hard-label}, our approach consistently outperforms state-of-the-art methods in terms of ASR. The results are the average of three random seeds. %FGSM's lower ASR is due to its single-step nature, leading to less effective adversarial samples. BIM, as an iterative refinement of FGSM, improves ASR by applying smaller, iterative perturbations. PGD, known for its high ASR, incorporates random initialization and a projection mechanism that ensures perturbations stay within the designated epsilon boundary, facilitating a more thorough exploration of adversarial spaces. Targeted attacks are generally more challenging than untargeted ones, as they require precise manipulations to mislead the target model into misclassifying an input as a specific incorrect category. 
FGSM's lower ASR stems from its single-step approach, while iterative methods like BIM and PGD achieve higher ASR by refining perturbations and exploring adversarial spaces more thoroughly. Targeted attacks are particularly challenging as they demand precise manipulations to force misclassification into a specific category.
%FGSM's lower ASR arises from its single-step approach, while BIM and PGD improve ASR through iterative refinement and random initialization, respectively, enabling more effective adversarial exploration. PGD's projection mechanism ensures perturbations remain within the epsilon boundary. Targeted attacks are inherently more difficult than untargeted ones, as they demand precise manipulations to induce specific misclassifications.
Our method consistently surpasses all baselines across the datasets detailed in \Cref{tab:asr-hard-label}, achieving a remarkable ASR of 70.79\% on the CIFAR-10 dataset. This performance significantly improves upon the best baseline, TEBA, which attains an ASR of 55.43\% under a $150k$ query budget. Additionally, on CIFAR-100, our method surpasses the previous best by 15.36\%. Further details in the soft-label setting are provided in (Appendix~\ref{sec:dfta_soft}) for comprehensive analysis. Since DifAttack~\cite{liu2024difattack} uses real data in a different setting from ours, and DERD~\cite{zhou2024derd}  lacks query efficiency and does not release its code, a direct comparison is not feasible. We also demonstrate in (Appendix~\ref{sec:dfta_prove}) that the use of synthetic data combined with the trained substitute model enhances transferability and significantly improves the ASR. Furthermore, we explore the relationship between query budget, model efficacy, and attack success in (Appendix~\ref{sec:s_one}) and (Appendix~\ref{sec:s_two}).

\section{Conclusion}
In this work, we identified that attackers could exploit diffusion model APIs to generate high-resolution, diverse synthetic images for training substitute models to facilitate model extraction and adversarial transfer attacks. We developed a method for training robust substitute models in a data-free, hard-label, and query-limited setting. Our API-based approach significantly reduced the query budget needed for effective model extraction and transfer attacks. Our experiments across seven datasets demonstrated that our method enhanced attack efficiency while minimizing the resources required to compromise target models.

%In this paper, we highlight a novel security risk posed by the misuse of text-to-image foundation models in black-box systems: an attacker can exploit diffusion model APIs to obtain high-resolution and diverse synthetic images for training a substitute model. To address this security risk, we tackled the task of training a robust substitute model from black-box target models in a data-free, hard-label, and query-limited setting. Previous studies have attempted similar model extraction and transfer attacks using this setting but faced issues with GAN-based frameworks such as convergence failures and model collapse, leading to inefficiency and extensive querying of black-box models. This made such attacks less threatening under real-world conditions. We revisit the relationship between the efficiency of the substitute model and the reduced query budget, designing an API-based attacking framework. Through experiments across seven datasets, our method significantly reduces the query budget needed to effectively access the target model. 
\newpage
\bibliographystyle{IEEEbib}
\bibliography{main}
\newpage
% %\input{AnonymousSubmission/ Checklist}
% \newpage

\appendix
\newpage
\section{Appendix}
\subsubsection{A. Query Budget and Training Settings for Data-Free Model Extraction}
\label{sec:dfme_setting}
During the data generation phase, we employ the Stable Diffusion model to create high-resolution images of 512x512 pixels, operating on the prompts we provided for 50 inference cycles. Once generated, we downscaled these images to align with the dimensions of the target datasets. For CIFAR-10, this meant scaling down from 512x512 to 32x32 pixels, and for ImageNet subsets, from 512x512 to 256x256 pixels. We produced a substantial collection of $200k$ synthetic samples per dataset, which served as the initial training material for our substitute models. To ensure consistency, the same training settings were used for both the original teacher models and the synthetic substitute models. This entailed using the SGD optimizer with a momentum of 0.9, a weight decay set at 
$5\times 10^{-4}$ , and a cosine learning rate scheduler commencing at a learning rate of 0.1. Our experiments conduct on an Nvidia-A40 GPU, and the entire distillation training process take only 10 minutes under a query budget of $5k$.

\textbf{Query Budget}. Our approach emphasizes query efficiency, so we consistently apply the same query constraint across all baseline methods. Specifically, the query budget is set to $Q=5k$ for CIFAR-10, $Q=1k$ for ImageNetette, $Q=130$ for ImageFruit, $Q=50$ for ImageYellow, and $Q=30$ for ImageSquaw. In the distillation phase, each teacher model undergoes a single forward pass with the allotted Q queries using synthetic data, which is the sole instance we query the teacher model. The top-1 hard-labels gleaned from this query are preserved for subsequent use. We then leverage the substitute model’s logits alongside the hard-labels from the teacher to compute the cross-entropy loss, thus advancing the substitute model's training. We measure performance by conducting three independent trials with 3 random seeds and present the mean top-1 accuracy.

\textbf{Data and Model}. We evaluate our method on a range of datasets to assess its performance under diverse conditions: CIFAR-10~\cite{krizhevsky2009learning}, CIFAR-100~\cite{krizhevsky2009learning}. In addition, we use specialized ImageNet subsets: ImageFruit~\cite{deng2009imagenet}, ImageYellow~\cite{deng2009imagenet}, and ImageSquaw~\cite{deng2009imagenet}, and Tiny ImageNet~\cite{le2015tiny}. To validate the generalization and practicality of our method, we experiment with different model architectures as the target model, including AlexNet~\cite{krizhevsky2012imagenet}, VGG-16~\cite{simonyan2014very}, VGG-19~\cite{simonyan2014very}, Wide-ResNet-16~\cite{zagoruyko2016wide}, and ResNet34~\cite{he2016deep}. Simultaneously, we also explore different model architectures as the substitute model, specifically VGG-16, VGG-19, Wide-ResNet-16, ResNet-18~\cite{he2016deep}, and ResNet-34 (see details in~\Cref{tab:diff_sub}). All the training parameters for the substitute model follow the settings in~\cite{truong2021data}. The target model, trained on private, real-world datasets, functions as a black-box and is only accessible to attackers via queries. Conversely, the substitute model is exclusively trained on synthetic data. This setup aims to evaluate the practicality of our method in environments where direct access to the target model's training data is restricted, effectively simulating a real-world adversarial scenario where attackers depend on synthetic approximations to challenge and compromise black-box models.

\textbf{Baselines}. In our experiments, we select two distinct categories of prevalent approaches. MAZE~\cite{kariyappa2021maze}, DFME~\cite{truong2021data}, and ZSDB3~\cite{wang2021zero} originally design for soft-label settings (i.e., probabilities or logits). Furthermore, we evaluate models designed for hard-label settings (i.e., return top-1 prediction), including DFMS~\cite{sanyal2022towards} and DisGuide~\cite{rosenthal2023disguide}.
\subsubsection{B. Performance Comparison by Soft-Label in Model Extraction}
\label{sec:soft_me}
\begin{table}[htbp]
\centering
%\vspace{-4mm}
\caption{Accuracy (\%) of substitute models on datasets with CIFAR-10, and ImageNet subsets in the soft-label setting. All results are averaged over three random seeds.}
\label{tab:t1}
\resizebox{0.48\textwidth}{!}{
\begin{tabular}{@{} c  c  c  c  c  c  c  c  c @{}}
\toprule

%Dataset  & Teacher & MAZE~\cite{alexey2016adversarial} & DFME~\cite{truong2021data} & ZSDB3~\cite{wang2021zero} & DFMS-HL~\cite{sanyal2022towards} & DisGuide~\cite{rosenthal2023disguide}   & Ours   & Query Budget \\ 
Dataset  & Teacher & MAZE & DFME & ZSDB3 & DFMS-HL & DisGuide   & Ours  & Query Budget \\ 

\midrule
CIFAR-10         & 93.9 & 10.7 & 10.8 & 11.1 & 11.4 & 13.4       & \textbf{88.8}                    & $5k$             \\
ImageNette       & 92.2 & 11.2 & 10.3 & 10.5 & 11.2 & 11.8           & \textbf{83.9}                    & $1k$             \\
ImageSquawk  & 92.4 & 10.4 & 10.6 & 10.2 & 11.2 & 11.7         & \textbf{83.6}                   & 30             \\
ImageFruit   & 78.2 & 10.2 & 10.4 & 11.1 & 10.9 & 11.3      & \textbf{70.8}                    & 130            \\
ImageYellow   & 90.8 & 10.2 & 10.4 & 10.6 & 11.7 & 11.8        & \textbf{82.4}                    & 50             \\
\bottomrule
\end{tabular}
}
\end{table}

\begin{table}[htbp]
\centering
%\vspace{-4mm}
\caption{Accuracy (\%) of student models on datasets of hundreds of classes in the soft-label setting. All results are averaged over three random seeds.}
\label{tab:t2}
\resizebox{0.47\textwidth}{!}{
\begin{tabular}{@{} c  c  c  c  c  c  c  c  c @{}}
\toprule
%Dataset  & Teacher & MAZE~\cite{alexey2016adversarial} & DFME~\cite{truong2021data} & ZSDB3~\cite{wang2021zero} & DFMS-HL~\cite{sanyal2022towards} & DisGuide~\cite{rosenthal2023disguide}   & Ours  & Query Budget \\ 
Dataset  & Teacher & MAZE & DFME & ZSDB3 & DFMS-HL & DisGuide   & Ours  & Query Budget \\ 
\midrule
CIFAR-100         & 79.89 & 1.05 & 1.19  &1.25  & 1.42 & 1.75     & \textbf{72.3}                     & $150k$             \\
Tiny-ImageNet        & 64.55 & 0.53 & 0.61  &0.67  & 0.72 & 0.86         & \textbf{58.5}                     & $200k$             \\

\bottomrule
\end{tabular}
}
\end{table} 
Our method achieves top-one accuracy rates of 88.9\% and 83.9\% on the CIFAR-10 and Imagenette datasets, respectively, marking a substantial improvement over all previous methods under a constrained query budget. This enhancement stems from the soft label setting during the distillation learning phase, wherein the substitute model accesses the target model's soft labels (i.e., probabilities or logits). This access enables the substitute to learn a richer set of information compared to the hard-label setting. As a result, both our method and the baseline demonstrate improvements over results obtained under hard-label conditions. However, the baseline's performance remains suboptimal, as demonstrated in~\Cref{fig:visualization}. Specifically, the GANs training framework, operating under a severely limited query budget, produces only noisy data, and the training loss fails to converge.
\subsubsection{Empirical Studies of Previous Methods}
\label{sec:exprical_loss}
To better demonstrate the training deficiencies of state-of-the-art methods DFME~\cite{truong2021data}, DisGuid~\cite{rosenthal2023disguide} under a restricted $5k$ query budget, this section provides an empirical analysis of the loss changes in their generator and substitute models. Our experiments are conducted on the CIFAR-10 dataset with a stringent limit of 20 epochs due to the $5k$ query budget, highlighting the challenges traditional GAN-based training methods face under such constraints.
\begin{figure}[htbp]
%\vspace{-4mm}
% \end{wrapfigure}
% \begin{figure}[t]
     \centering
     \includegraphics[width=0.46\textwidth]{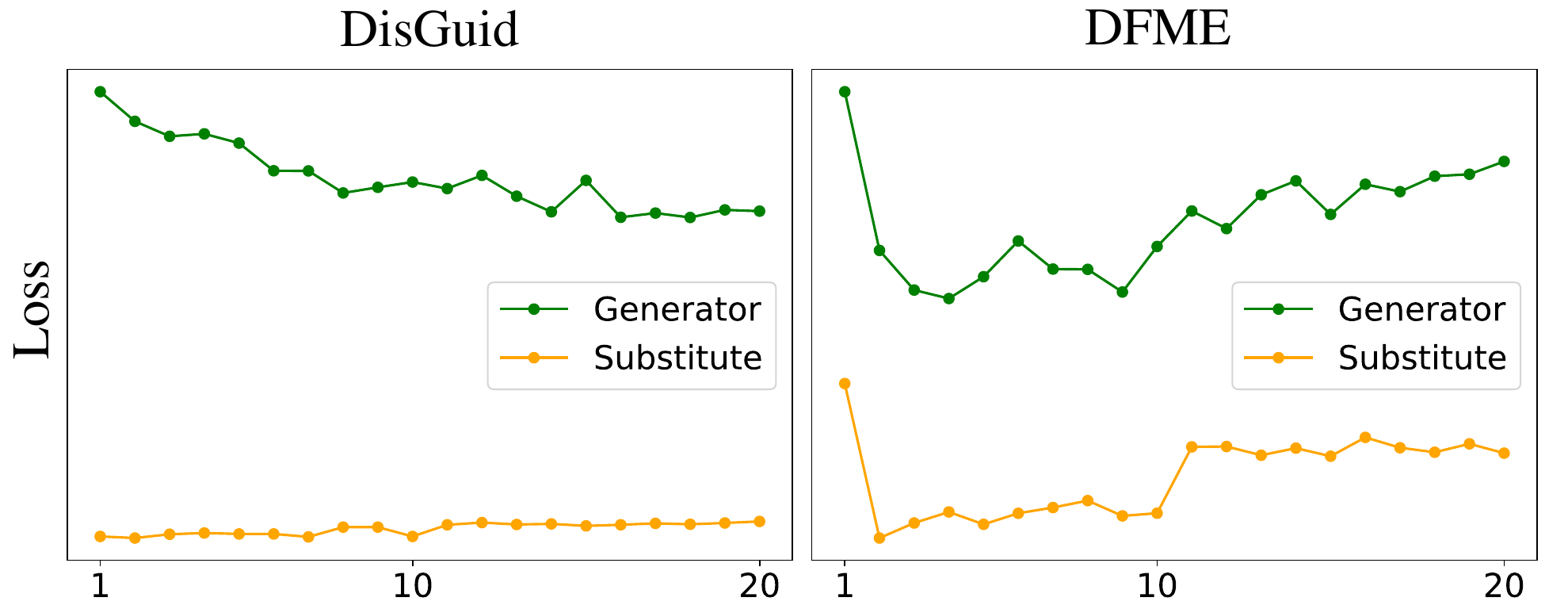}
     \caption{Training flaws of previous SOTA methods under $5k$ query budget.
% achieve better 
% diversity, clear patterns can be observed (\eg 0 and 8).
}
     \label{fig:visualization}
     %\vspace{-6mm}
% \end{figure}
\end{figure}
As depicted in~\Cref{fig:visualization}, the left subplot for DisGuid shows a slight reduction in generator loss (green), decreasing from an initial 5.5 to 4.8 before it begins to oscillate. However, the loss does not converge and remains high, indicating that the generator fails to produce high-quality images effectively. The loss for the substitute model (orange) remains near zero, suggesting that the generator struggles to generate consistently challenging images, thereby hindering the substitute model's ability to learn effectively from the target model. The right subplot in~\Cref{fig:visualization} presents the loss values for DFME, where both the generator and substitute model exhibit significant fluctuations throughout the training process. This unstable convergence pattern indicates that under a minimal query budget, the substitute model struggles to match the target model’s output, further highlighting the inherent challenges of training traditional GANs to converge under these conditions.
\subsubsection{Ablation Study on Pre-trained Models}
\label{sec:abl}
In this section, we assess the impact of the pre-training stage on the performance of substitute models. As indicated in \Cref{tb:md}, eliminating the pre-training stage leads to significant performance degradation. Specifically, we evaluated our method on CIFAR-10, an ImageNet subset, CIFAR-100, and Tiny ImageNet, using the same query budgets outlined in (Appendix A). The observed degradations in test accuracy on these datasets were 33.9\%, 16.4\%, 17.1\%, and 9.1\%, respectively. Our training framework diverges from traditional adversarial training of generators and substitutes by leverage the online stable diffusion APIs to synthesize high-quality images. This approach is critical for reducing query budgets, as pre-training the substitute model proves essential. An effectively pre-trained substitute model not only reduces the number of queries required but also accelerates and enhances the efficiency of the training convergence process during the knowledge distillation stage.

\begin{table}[htpb]
%\vspace{-4mm}
\centering
\caption{Ablation studies evaluating the impact of the pre-training stage with Top-1 accuracy. All results are averaged over three random seeds.}
%\jz{w/ and w/o}
\label{tb:md}
\scalebox{0.65}{
\begin{tabular}{ccccc}
\toprule
Method & CIFAR-10 & ImageNet subset & CIFAR-100 & Tiny-ImageNet \\
\midrule
w/ pre-training stage & \textbf{81.8} & \textbf{69.7} & \textbf{60.5} & \textbf{45.9} \\
w/o pre-training stage & 47.9 & 53.3 & 43.4 & 36.8 \\
\bottomrule
\end{tabular}}
\vspace{-4mm}
\end{table}
\subsubsection{E. Query Budget and Evaluations For Data-Free Adversarial Transfer Attack}
\label{sec:dfta_setting}
Adversarial transfer presents a more complex challenge than model extraction because it involves creating adversarial samples that must maintain high transferability between the substitute and target models. Consequently, we allocate a query budget of $150k$ for both the CIFAR-10 and CIFAR-100 datasets to accommodate this complexity.  Our experiments conduct on an Nvidia-A40 GPU, and the entire adversarial attacking process take about 45 minutes under a query budget of $150k$. To ensure fair comparisons, the same query budget is applied to the baselines.
We utilize well-known adversarial attack methods such as BIM~\cite{kurakin2016adversarial}, FGSM~\cite{alexey2016adversarial}, PGD~\cite{madry2017towards} for conducting experiments. The specific parameters set for these experiments on CIFAR-10 and CIFAR-100 include a perturbation limit $\epsilon$ = 8/255 and the step size at $\alpha$ = 2/255 follow by the setting in~\cite{zhang2022towards}. In the untargeted attack mode, adversarial examples are generated only from images that the model initially classifies correctly. In contrast, targeted attack strategies generate adversarial examples solely from images that are not already misclassified into specific incorrect categories. The attack success rate is calculated using the ratio $n/m$, where $n$ is the number of adversarial examples that successfully fool the attacked model, and $m$ is the total number of adversarial examples created.

\textbf{Baselines}. We select the most prominent baselines for black-box adversarial attacks, including JPBA~\cite{papernot2017practical} and Knockoff~\cite{orekondy2019knockoff}, which require access to training data. Additionally, we evaluate black-box knowledge distillation methods exploiting probabilities returned by the target model, with a focus on DFME~\cite{truong2021data}. Furthermore, we critically examine data-free black-box attacks using hard-labels, closely align with our experimental setup, as detailed in (Appendix E), including DaST~\cite{zhou2020dast} and TEBA~\cite{zhang2022towards}, to underscore the comparative effectiveness of our approach.
\subsubsection{F. Impact Analysis of Query Efficiency in  Adversarial Transfer Attack}
\label{sec:dfta_prove}
An adversarial example \( x' \) for the model \( f_{\text{sub}} \) of the substitute is generated by perturbing \( x \) such that \( x' = x + \delta \), where \( \delta \) is chosen to maximize \( L_{\text{sub}}(x', y; \theta) \). The goal of the adversarial attack is to find \( x' \) such that:
\[
\arg \max f_{\text{sub}}(x') \neq y,
\]
and ideally:
\[
\arg \max f_{\text{target}}(x') \neq y,
\]
The transferability of adversarial examples from \( f_{\text{sub}} \) to \( f_{\text{target}} \) can be quantified as:

% \begin{align*}
% T(f_{\text{sub}} \rightarrow f_{\text{target}}) 
% &= \\
% &\mathbb{P}(\arg \max f_{\text{target}}(x + \delta) \neq y \mid \arg \max f_{\text{sub}}(x + \delta) \neq y)
% \end{align*}

\begin{align*}
T(f_{\text{sub}} \rightarrow f_{\text{target}}) 
&= 
\mathbb{P}(\arg \max f_{\text{target}}(x + \delta) \neq y \mid \\
&\quad \arg \max f_{\text{sub}}(x + \delta) \neq y),
\end{align*}

\subsubsection*{Training on Synthetic Data.}

Objective for non-pretrained substitute model \( f_{\text{sub}}^{\text{real}} \):
\[
f_{\text{sub}}^{\text{real}} = \min_{f_{\text{sub}}} \mathbb{E}_{x, y \sim \mathcal{D}_{\text{real}}} [L(f_{\text{sub}}(x), y)],
\]

Objective for pretrained substitute model \( f_{\text{sub}}^{\text{syn}} \):
\[
f_{\text{sub}}^{\text{syn}} = \min_{f_{\text{sub}}} \mathbb{E}_{x, y \sim \mathcal{D}_{\text{syn}}} [L(f_{\text{sub}}(x), y)],
\]

\subsubsection*{Effect on Decision Boundary.}
The decision boundary in \( f_{\text{sub}}^{\text{syn}} \) tends to be broader because the diffusion model generates diverse training data. We define the decision boundary of a model \( f \) as \( \partial f = \{ x \mid \arg \max f(x) \text{ changes} \} \). If the decision boundary \( \partial f_{\text{syn}} \) of the substitute model is close to that of the target model \( \partial f_{\text{target}} \), then the adversarial samples generated on \( f_{\text{sub}}^{\text{syn}} \) are more likely to transfer effectively to \( f_{\text{target}} \).

% \begin{itemize}
%     \item Broader decision boundary in \( f_{\text{sub}}^{\text{syn}} \) because the diffusion model generates diverse training data.
%     \item Denote the decision boundary of \( f \) as \( \partial f = \{ x \mid \arg \max f(x) \text{ changes} \} \).
%     \item If \( \partial f_{\text{syn}} \) is close to \( \partial f_{\text{target}} \), the adversarial samples generated on \( f_{\text{sub}}^{\text{syn}} \) are likely to transfer well to \( f_{\text{target}} \).
% \end{itemize}

\subsubsection*{Query Efficiency.}

    Let \( Q(f_{\text{sub}} \rightarrow f_{\text{target}}) \) represent the number of queries required to generate a successful adversarial example for the target using \( f_{\text{sub}} \).
    For \( f_{\text{sub}}^{\text{real}} \), the expected number of queries can be denoted as:
    \[
    \mathbb{E}[Q(f_{\text{sub}}^{\text{real}} \rightarrow f_{\text{target}})] = \frac{1}{T(f_{\text{sub}}^{\text{real}} \rightarrow f_{\text{target}})}  ,
    \]
    For \( f_{\text{sub}}^{\text{syn}} \), due to reduced transferability \( T(f_{\text{sub}}^{\text{syn}} \rightarrow f_{\text{target}}) \), the expected number of queries decreased:
    \[
    \mathbb{E}[Q(f_{\text{sub}}^{\text{syn}} \rightarrow f_{\text{target}})] = \frac{1}{T(f_{\text{sub}}^{\text{syn}} \rightarrow f_{\text{target}})}  ,
    \]
    Since \( T(f_{\text{sub}}^{\text{real}} \rightarrow f_{\text{target}}) \leq T(f_{\text{sub}}^{\text{syn}} \rightarrow f_{\text{target}}) \) implies:
    \[
    \mathbb{E}[Q(f_{\text{sub}}^{\text{syn}} \rightarrow f_{\text{target}})] \leq \mathbb{E}[Q(f_{\text{sub}}^{\text{real}} \rightarrow f_{\text{target}})].
    \]
\subsubsection*{Conclusion}
Pretraining the substitute model requires fewer queries to achieve an adversarial transfer attack.

\subsubsection{G. Interplay Between Substitute Model Training, Model Efficacy, and Attack Success}
\label{sec:s_one}
\begin{figure}[t]
%\vspace{-1mm}
     \centering
     \includegraphics[scale=0.258]{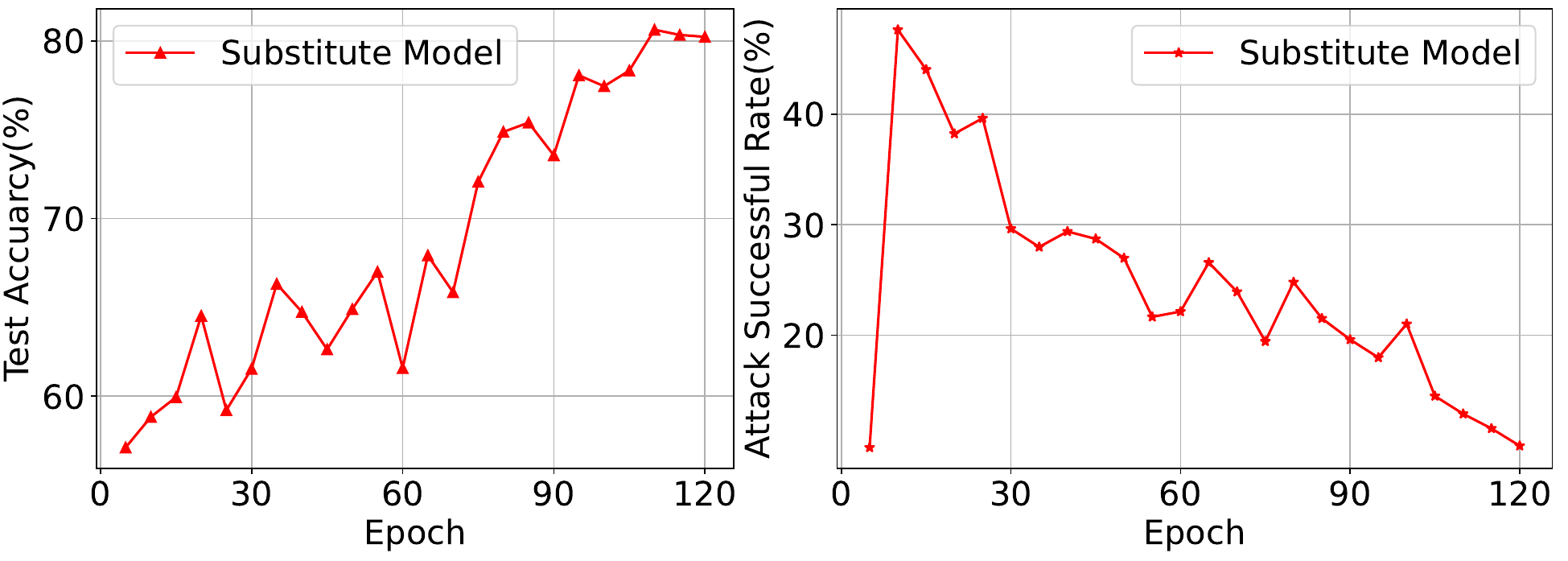}
     \caption{\textbf{Left}: Changes in ACC with different checkpoint in stage one.
            \textbf{Right}: Changes in ASR with different checkpoint in stage one.
}
     \label{fig:stage1}
     %\vspace{-4mm}
\end{figure}

In the initial stage of our experiments, we leverage off-the-shelf generative models to synthesize high-quality synthetic data based on benign prompts, followed by the pre-training of the substitute model. Our objective was to assess the changes in test accuracy and ASR across different training epochs. Specifically, our experiments were conducted on the CIFAR-10 dataset using a ResNet-18 substitute model. We saved a checkpoint every five epochs and subsequently evaluated each checkpoint for its corresponding test ACC and ASR.The results, as depicted in \Cref{fig:stage1}, demonstrate that: 
1) the test accuracy progressively increases with the advancement of the pre-training process, and the checkpoint with the highest accuracy indeed provides an excellent starting point for the distillation training in the subsequent stage. 2) The experiments indicate that the ASR initially rises rapidly within the first 0-5 epochs and then gradually decreases. Consequently, if users require a model with a higher ASR, early stopping is recommended to capture a model that delivers superior performance.

\subsubsection{Analyzing the Impact of Query Budget on Substitute Model Efficacy and Attack Success}
\label{sec:s_two}
\begin{figure}[htbp]
%\vspace{-1mm}

     \centering
     \includegraphics[scale=0.26]
     {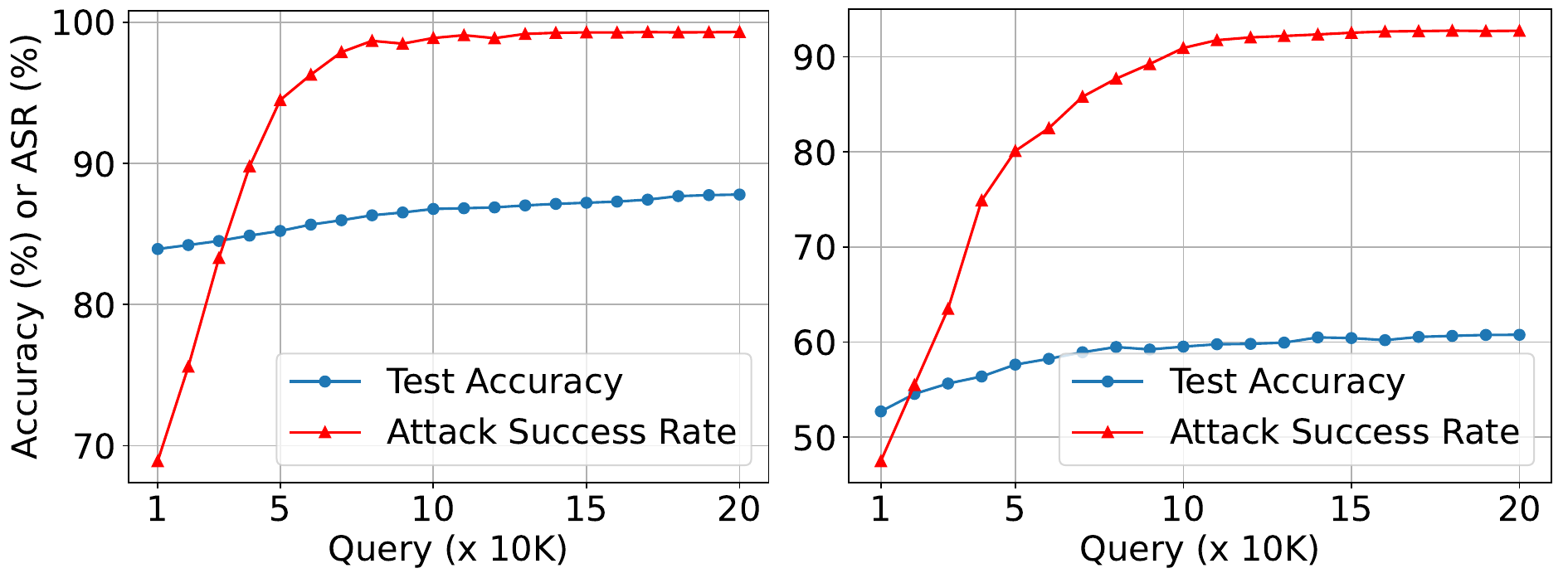}
     %\vspace{-1mm}
     \caption{\textbf{Left}: ACC and ASR on CIFAR-10 with different query budget.
            \textbf{Right}: ACC and ASR on CIFAR-100 with different query budget.
            %\jz{font size is too small!}
            }
  \label{fig:S2_acc_asr}
     %\vspace{-4mm}
\end{figure}

%\vspace{-2mm}
In this study, we investigate the relationship between two attack methods and the query budget, as well as the interrelation between these attacks. As illustrated in~\Cref{fig:S2_acc_asr}, the top-one accuracy of the substitute model improved from 84.22\% to 87.8\% as the query budget increased from $10k$ to $200k$. For adversarial transfer attacks employing untargeted and BIM methods, the ASR rose from 68.79\% to 99.32\%, demonstrating convergence. This enhancement can be attributed to the use of diverse and high-fidelity images generated by off-the-shelf generative models, which pre-train the substitute model and provide a rich knowledge base. This approach significantly enhances the ability to produce more transferable adversarial examples during attacks. Furthermore, we observed a positive correlation between the test accuracy of the substitute model and its ASR, confirming that the performance of data-free model extraction and data-free transfer attack tasks is positively correlated.
\subsubsection{Performance Comparison by Soft-Label in Adversarial Attack}
\label{sec:dfta_soft}
In this section, recognizing the suboptimal performance of previous baselines in the hard-label setting, we compare their effectiveness in the more favorable soft-label setting for a fairer and more comprehensive evaluation. Both our method and the baselines show improved performance under soft-label conditions, as this setting allows the substitute model to capture more knowledge. However, the baseline’s performance remains suboptimal. Specifically, our method achieves an ASR of 98.68\% on CIFAR-10 and 99.68\% on CIFAR-100 as shown in~\Cref{tab:asr-soft-label}, reflecting improvements of 41.85\% and 41.74\%, respectively, over the previous baselines.

\begin{table}[t]
\centering
\caption{ASR(\%) comparisons between our proposed method and baselines over CIFAR-10 and CIFAR-100 under soft-label settings with a query budget Q = $150k$. Best result in bold. All results are averaged over three random seeds.}
\label{tab:asr-soft-label}
\scalebox{0.78}{
\begin{tabular}{c c ccc ccc}
\toprule
\multirow{2}{*}{{Dataset}} & Type &
\multicolumn{3}{c}{ \small{}Targeted, soft-label}  & \multicolumn{3}{c}{ \small{}Untargeted, soft-label} \tabularnewline
\cmidrule(lr){2-8}
 & {\small{}Method} & {\small{}FGSM} & {\small{}BIM} & {\small{}PGD} & {\small{}FGSM} & {\small{}BIM} & {\small{}PGD} \tabularnewline

\midrule 

\multirow{6}{*}{{\small{}CIFAR-10}}
& {\small{}JPBA} & {\small 2.74} & {\small 3.86} & {\small 3.93} & {\small 8.29} & {\small 10.92} & {\small 8.62} \tabularnewline

& {\small{}Knockoff} & {\small 2.16} & {\small 3.53} & {\small 3.37} & {\small 7.55} & {\small 10.18} & {\small 9.06} \tabularnewline

& {\small{}DaST} & {\small 3.95} & {\small 4.19} & {\small 4.33} & {\small 8.98} & {\small 12.53} & {\small 8.32} \tabularnewline

& {\small{}DFME} & {\small 3.55} & {\small 11.28} & {\small 8.93} & {\small 15.13} & {\small 20.17} & {\small 17.89} \tabularnewline

& {\small{}TEBA} & {\small 10.38} & {\small 31.8} & {\small 27.9} & {\small 34.48} & {\small 56.83} & {\small 50.72} \tabularnewline

\cmidrule(lr){2-8}
& {\small{}\textbf{Ours}} & {\small \textbf{17.27}} & {\small \textbf{88.71}} & {\small \textbf{85.85}} & {\small \textbf{63.95}} & {\small \textbf{98.68}} & {\small \textbf{98.63}} \tabularnewline

\midrule 

\multirow{6}{*}{{\small{}CIFAR-100}}
& {\small{}JPBA} & {\small 3.25} & {\small 4.19} & {\small 4.24} & {\small 9.39} & {\small 11.45} & {\small 9.75} \tabularnewline

& {\small{}Knockoff} & {\small 2.67} & {\small 4.02} & {\small 3.87} & {\small 8.55} & {\small 11.28} & {\small 10.13} \tabularnewline

& {\small{}DaST} & {\small 4.45} & {\small 4.69} & {\small 4.83} & {\small 9.98} & {\small 13.53} & {\small 9.92} \tabularnewline

& {\small{}DFME} & {\small 4.55} & {\small 12.28} & {\small 9.93} & {\small 16.13} & {\small 21.17} & {\small 18.89} \tabularnewline

& {\small{}TEBA} & {\small 11.38} & {\small 32.8} & {\small 28.9} & {\small 35.48} & {\small 57.94} & {\small 51.72} \tabularnewline

\cmidrule(lr){2-8}
& {\small{}\textbf{Ours}} & {\small \textbf{18.27}} & {\small \textbf{89.71}} & {\small \textbf{86.85}} & {\small \textbf{64.95}} & {\small \textbf{99.68}} & {\small \textbf{99.63}} \tabularnewline

\bottomrule
\end{tabular}}
\vspace{-2mm}
\end{table}
\subsubsection{Comparisons with Different Substitute Model}
\label{sec:different_sub}
%In this section, we aim to demonstrate the generalization and practicality of our method in~\Cref{tab:diff_sub}. We seek to prove that an attacker can employ various model architectures to perform model extraction attacks on black-box systems. We fix the target model architecture as ResNet-34 and conducted attacks using heterogeneous architectures, such as VGG16 and VGG19, as well as homogeneous architectures, including ResNet-18, ResNet-34, and Wide-ResNet-16-8. The experimental results show that, within our attack framework, all of the substitute models mentioned above achieve superior performance compared to previous baselines. For instance, our method improves the attack success rate by approximately 68\% and 56\% on CIFAR-10 and CIFAR-100 when using VGG16 as the substitute model, compared to prior baselines.
In this section, we demonstrate the generalization and practicality of our method as shown in~\Cref{tab:diff_sub}. We prove that an attacker can use various model architectures to perform model extraction attacks on black-box systems. Fixing the target model as ResNet-34, we conducted attacks using both heterogeneous architectures (VGG16, VGG19) and homogeneous ones (ResNet-18, ResNet-34, Wide-ResNet-16-8). The results show that all substitute models in our framework outperform previous baselines. Notably, using VGG16 as the substitute model, our method improves the attack success rate by 68\% on CIFAR-10 and 56\% on CIFAR-100 compared to prior baselines.

\begin{table}
\centering
\caption{Top-1 accuracy comparison between our method and previous baselines using different substitute model architectures. To evaluate the generalization of our approach, the target model was fixed as ResNet-34, while VGG-16 (V-16), VGG-19 (V-19), Wide-ResNet-16-8 (WRN16), ResNet-18 (R-18), and ResNet-34 (R-34) were employed as substitute models in the hard-label setting. CIFAR-10 and CIFAR-100 employ query budgets of 5k and 150k, respectively. All results are averaged over three random seeds.}
\label{tab:diff_sub}
\scalebox{0.73}{
\begin{tabular}{c c ccccc}
\toprule
\multirow{1}{*}{Dataset} & \multirow{1}{*}{Method} & 
V-16 & V-19 & WRN16 & R-18 & R-34  \tabularnewline
\midrule
\multirow{4}{*}{CIFAR-10}
 & DFME & 10.52 & 10.21 & 10.36 & 10.96 & 10.85 \tabularnewline
 & ZSDB3 & 10.43 & 10.31 & 10.20 & 10.83 & 10.47  \tabularnewline
 & DisGuide & 12.49 & 11.20 & 12.12 & 12.57 & 13.37 \tabularnewline
 \cmidrule(lr){2-7}
 & Ours & 80.73 & 80.44 & 83.74 & 81.51 & 84.41 \tabularnewline
 \midrule
\multirow{4}{*}{CIFAR-100}
 & DFME& 1.06 & 1.08 & 1.05 & 1.05 & 1.07  \tabularnewline
 & ZSDB3 & 1.03 & 1.05 & 1.07 & 1.05 & 1.09  \tabularnewline
 & DisGuide & 1.13 & 1.27 & 1.19 & 1.27 & 1.30  \tabularnewline
 \cmidrule(lr){2-7}
 & Ours & 57.56 & 58.86 & 59.14 & 60.51 & 58.82 \tabularnewline
\bottomrule
\end{tabular}}
\vspace{-6mm}
\end{table}
\subsubsection{Visualization of Synthetic Data} 
%\vspace{-4mm}
% begin{figure}[t]
% %\vspace{-1mm}
% \end{wrapfigure}
% \begin{figure}[t]
%     \centering
%     \includegraphics[scale=0.26]{img/vis_baseline.pdf}
%     \caption{Visualization of synthetic images generated by baseline DFME and Disguide with a $5k$ query budget.%\jz{font size is too small!}
% achieve better 
% diversity, clear patterns can be observed (\eg 0 and 8).
% }
% \vspace{-2mm}
% \label{fig:vis_baseline}
% \end{figure}
\begin{figure}[t]
    \centering
    \includegraphics[scale=0.26]{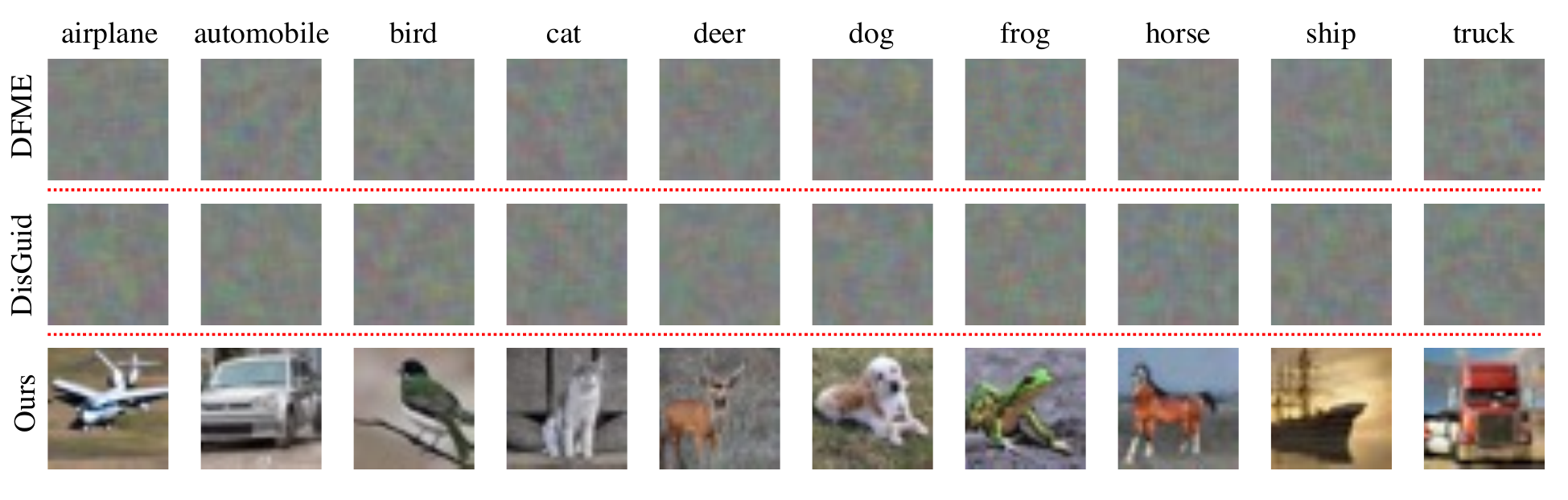}
    \caption{Visualization of synthetic images generated by baseline DFME and DisGuide with a $5k$ query budget.}
    \vspace{-2mm}
    \label{fig:vis_baseline}
\end{figure}

In this subsection, we present synthesized examples from DFME~\cite{truong2021data}, DisGuide~\cite{rosenthal2023disguide}, and our method to investigate the underlying reasons for the baseline performance approximating random guessing. Our objective is to corroborate the experimental results shown in~\Cref{tab:t1_hard} and~\Cref{tab:t2_hard} by demonstrating the synthesized images and training stage loss (see details Appendix~\ref{sec:exprical_loss}) produced by DFME and DisGuide within a $5k$ query budget constraint. As illustrated in~\Cref{fig:vis_baseline}, the images generated by DFME and DisGuide under this limited query budget appear nearly as random noise, lacking discernible patterns. This similarity in noisy patterns indicates that these GAN-based generators struggle with training under data scarcity, leading to model collapse and difficulties in achieving convergence.
%\vspace{-2mm}
\begin{table}[htpb]
%\vspace{-4mm}
\centering
\caption{Ablation studies evaluating the impact of the pre-training stage with Top-1 accuracy. All results are averaged over three random seeds.}
%\jz{w/ and w/o}
\label{tb:abl}
\scalebox{0.65}{
\begin{tabular}{ccccc}
\toprule
Method & CIFAR-10 & ImageNet subset & CIFAR-100 & Tiny-ImageNet \\
\midrule
w/ pre-training stage & \textbf{81.8} & \textbf{69.7} & \textbf{60.5} & \textbf{45.9} \\
w/o pre-training stage & 47.9 & 53.3 & 43.4 & 36.8 \\
\bottomrule
\end{tabular}}
\vspace{-4mm}
\end{table}
\subsubsection{Effect Investigation of Pre-training model in Model Extraction}
\label{sec:dfme_prove}
In model extraction, the objective is to approximate the target model's behavior by training a substitute model on the target model's outputs. The parameters of the substitute model, denoted as \( \theta_{\text{sub}} \), are optimized using the cross-entropy loss function, which is minimized through gradient-based updates. Given an input \( x \) and its corresponding one-hot label \( y \), the predicted probability for class \( y_i \) by the substitute model is \( P_{\theta_{\text{sub}}}(y_i|x) \). The cross-entropy loss function is defined as follows: 
\begin{equation}
\vspace{-1mm}
L_{\text{CE}}(P_{\theta_{\text{sub}}}(y|x), y) = -\sum_{i=1}^{C} y_i \log P_{\theta_{\text{sub}}}(y_i|x),
\end{equation}
where \( C \) denotes the total number of classes, and \( y_i \) is the \( i \)-th component of the label \( y \), where \( y_i = 1 \) if \( y = i \) and \( y_i = 0 \) otherwise. This loss function quantifies the divergence between the predicted and true distributions. To optimize the substitute model parameters \( \theta_{\text{sub}} \), we compute the gradient of the loss function:
\begin{equation}
\nabla_{\theta_{\text{sub}}} L_{\text{CE}}(P_{\theta_{\text{sub}}}(y|x), y) = -\sum_{i=1}^{C} y_i \nabla_{\theta_{\text{sub}}} \log P_{\theta_{\text{sub}}}(y_i|x), 
\end{equation}
%Here, \( \nabla_{\theta_{\text{sub}}} \log P_{\theta_{\text{sub}}}(y_i|x) \) denotes the gradient of the log-probability function with respect to the model parameters \( \theta_{\text{sub}} \). Applying the chain rule, this gradient can be expanded as:
where \( \nabla_{\theta_{\text{sub}}} \log P_{\theta_{\text{sub}}}(y_i|x) \) denotes the gradient of the log-probability function with respect to \( \theta_{\text{sub}} \). Using the chain rule, this gradient expands to:
\begin{equation}
\nabla_{\theta_{\text{sub}}} \log P_{\theta_{\text{sub}}}(y_i|x) = \frac{1}{P_{\theta_{\text{sub}}}(y_i|x)} \nabla_{\theta_{\text{sub}}} P_{\theta_{\text{sub}}}(y_i|x),
\end{equation}
% where \( \nabla_{\theta_{\text{sub}}} P_{\theta_{\text{sub}}}(y_i|x) \) is the gradient of the predicted probability \( P_{\theta_{\text{sub}}}(y_i|x) \) with respect to the parameters \( \theta_{\text{sub}} \). Substituting this expression back into the original gradient formula, we obtain the final gradient expression:
where \( \nabla_{\theta_{\text{sub}}} P_{\theta_{\text{sub}}}(y_i|x) \) denotes the gradient of the predicted probability with respect to \( \theta_{\text{sub}} \). Substituting this into the original gradient formula yields the final expression:
\begin{equation}
\nabla_{\theta_{\text{sub}}} L_{\text{CE}}(P_{\theta_{\text{sub}}}(y|x), y) = -\sum_{i=1}^{C} \frac{y_i}{P_{\theta_{\text{sub}}}(y_i|x)} \nabla_{\theta_{\text{sub}}} P_{\theta_{\text{sub}}}(y_i|x),
\end{equation}
% This formula describes the gradient update method for the substitute model in the model extraction task. By iteratively updating the parameters \( \theta_{\text{sub}} \), the substitute model can increasingly approximate the target model's behavior, thereby achieving efficient model extraction.
This formula outlines the gradient update process for the substitute model in model extraction. Iterative updates to \( \theta_{\text{sub}} \) progressively align the substitute model's behavior with that of the target model, enabling efficient model extraction.
\subsubsection*{Case 1: Pretrain with Synthetic Images.} 
%In the first case, we pretrain the substitute model \( \theta_{\text{sub}}^{\text{pretrain}} \) using synthetic images. The objective is to minimize the expected cross-entropy loss between the predictions of the substitute model and the true labels:
We first pretrain the substitute model \( \theta_{\text{sub}}^{\text{pretrain}} \) on synthetic images, aiming to minimize the expected cross-entropy loss between the substitute model's predictions and the true labels:
\begin{equation}
%\vspace{-1mm}
\theta_{\text{sub}}^{\text{pretrain}} = \arg \min_{\theta_{\text{sub}}} \mathbb{E}_{x \sim \mathcal{D}_{\text{syn}}} \left[ L_{\text{CE}} \left( P_{\theta_{\text{sub}}}(y|x), y \right) \right],
\end{equation}
where \( \mathcal{D}_{\text{syn}} \) denotes the synthetic dataset, and \( L_{\text{CE}} \) is the cross-entropy loss. Given that the synthetic data distribution \( \mathcal{D}_{\text{syn}} \) is designed to approximate the real data distribution \( \mathcal{D}_{\text{real}} \), the pretrained parameters \( \theta_{\text{sub}}^{\text{pretrain}} \) are expected to be closely align to the target model's parameters \( \theta_{\text{target}} \).
\subsubsection*{Fine-tune Step.} 
%\vspace{-2mm}
After pre-training, we fine-tune the substitute model using a limited set of queries.
\begin{equation}
\theta_{\text{sub}}^{\text{fine-tune}} = \arg \min_{\theta_{\text{sub}}} \mathbb{E}_{x \sim \hat{\mathcal{X}}_{\text{query}}} \left[ L_{\text{CE}} \left( P_{\theta_{\text{sub}}}(y|x), P_{\theta_{\text{target}}}(y|x) \right) \right],
\end{equation}
where \( \hat{\mathcal{X}}_{\text{query}} \) represents the set of input queries sampled for fine-tuning. Given that \( \theta_{\text{sub}}^{\text{pretrain}} \) is already close align to \( \theta_{\text{target}} \), the gradient of the loss function with respect to the pretrained parameters is expected to be small:\begin{equation}
%\vspace{-1mm}
\nabla_{\theta_{\text{sub}}^{\text{pretrain}}} L_{\text{CE}} \left( P_{\theta_{\text{sub}}^{\text{pretrain}}}(y|x), P_{\theta_{\text{target}}}(y|x) \right) \approx 0,
\end{equation}
This small gradient implies that only minor adjustments are necessary during fine-tuning, making the process efficient and requiring fewer queries to the target model.
\subsubsection*{Case 2: Train from Scratch.}
In contrast, when training the substitute model from scratch, the optimization problem is formulated as:
\begin{equation}
\theta_{\text{sub}}^{\text{scratch}} = \arg \min_{\theta_{\text{sub}}} \mathbb{E}_{x \sim \hat{\mathcal{X}}_{\text{query}}} \left[ L_{\text{CE}} \left( P_{\theta_{\text{sub}}}(y|x), P_{\theta_{\text{target}}}(y|x) \right) \right],
\end{equation}
Since the parameters \( \theta_{\text{sub}}^{\text{scratch}} \) start from an uninitialized state, far from the target model's parameters \( \theta_{\text{target}} \), the initial gradients will be large:
\begin{equation}
\nabla_{\theta_{\text{sub}}^{\text{scratch}}} L_{\text{CE}} \left( P_{\theta_{\text{sub}}^{\text{scratch}}}(y|x), P_{\theta_{\text{target}}}(y|x) \right) \gg 0.
\end{equation}
\subsubsection*{Conclusion.} The comparison of these two scenarios illustrates that a synthetically pre-trained substitute model significantly reduces the need for extensive parameter adjustments, requiring fewer queries and leading to a more efficient extraction process, while training from scratch involves larger gradients and demands more queries to the target model, resulting in a longer training period.

\newpage
%\bibliographystyle{IEEEbib}
% \bibliography{main_ref}
%\newpage

%\bibliography{latex/appedx_ref}

\end{document}